\documentclass[letterpaper, 10 pt, journal, twoside]{IEEEtran} 
\usepackage[backend=biber,style=ieee,natbib=true,maxnames=3]{biblatex} 
\usepackage{hyperref}

\usepackage{amsmath}

\usepackage{algorithm}
\usepackage{algpseudocode}

\usepackage[caption=false,font=footnotesize]{subfig}

\usepackage{fancyhdr}
\usepackage{tabularx}
\usepackage{multirow}
\usepackage{graphicx}
\usepackage{cleveref}
\usepackage{booktabs}
\usepackage{color}

\newcommand{\gnnobj}{e(PG \cup l)}
\newcommand{\gnnobjest}{\tilde{e}(PG \cup l)}

\newcommand{\argmin}{\arg\!\min} 

\newif\ifshowrev

\newcommand*{\rev}[1]{
\ifshowrev
\leavevmode\unskip {\color{red}\textrm{#1}} \unskip
\else 
\leavevmode\unskip #1 \unskip
\fi 
}

\newif\ifshowrevfinal

\newcommand*{\revfinal}[1]{
\ifshowrevfinal
\leavevmode\unskip {\color{red}\textrm{#1}} \unskip
\else 
\leavevmode\unskip #1 \unskip
\fi 
}

\newif\ifshowral

\newcommand*{\ral}[1]{
\ifshowral
\leavevmode\unskip #1 \unskip
\else 
\fi 
}

\newif\ifshowarxiv

\newcommand*{\arxiv}[1]{
\ifshowarxiv
\leavevmode\unskip #1 \unskip
\else 
\fi 
}

\newcommand*{\ralheader}[1]{
\ifshowral
#1
\else 
\fi 
}
\newcommand*{\arxivheader}[1]{
\ifshowarxiv
#1
\else 
\fi 
}
\usepackage{csquotes}
\MakeOuterQuote{"}
\usepackage{gensymb}

\newcommand{\ph}[1]{{\textbf{#1:}}}
\usepackage{tikz}

\begin{document}

\showrevfalse
\showrevfinalfalse

\showralfalse
\showarxivtrue

%
\title{Loop Closure Prioritization for Efficient and Scalable Multi-Robot SLAM}

\ralheader{\markboth{IEEE Robotics and Automation Letters. Preprint Version. Accepted July, 2022}{Denniston \MakeLowercase{\textit{et al.}}: Loop Closure Prioritization}}

\arxivheader{
\fancypagestyle{firstpage}
{
    \fancyhead[C]{Accepted for publication at RA-L and IROS 2022, please cite as follows: \\
    C. E. Denniston, Y. Chang, A. Reinke, K. Ebadi, G. S. Sukhatme, L. Carlone, B. Morrell, A. Agha-mohammadi. \\ 
    "Loop Closure Prioritization for Efficient and Scalable Multi-Robot SLAM." IEEE Robot. and Autom. Letters (RA-L), 2022
    }
}
}

\author{Christopher E. Denniston\textsuperscript{\textdagger,1,3},
Yun Chang\textsuperscript{\textdagger,2 },
Andrzej Reinke\textsuperscript{3,4}, 
Kamak Ebadi\textsuperscript{3},\\
Gaurav S. Sukhatme\textsuperscript{1}, 
Luca Carlone\textsuperscript{2},
Benjamin Morrell\textsuperscript{3},
Ali-akbar Agha-mohammadi\textsuperscript{3}
\ral{\thanks{Manuscript received: February, 24, 2022; Revised May 29, 2022; Accepted June 24, 2022}}.
\thanks{This paper was recommended for publication by Editor Sven Behnke upon evaluation of the Associate Editor and Reviewers' comments.}
\thanks{\textsuperscript{\textdagger}Equal contribution. Corresponding Authors: {\tt\footnotesize cdennist@usc.edu, yunchang@mit.edu}. \textsuperscript{1}University of Southern California \textsuperscript{2}Massachusetts Institute of Technology \textsuperscript{3}Jet Propulsion Laboratory, California Institute of Technology \textsuperscript{4}University of Bonn, Germany. Sukhatme holds concurrent appointments as a Professor at USC and as an Amazon Scholar. His work on this paper was performed at USC and is not associated with Amazon. }
\thanks{This work was supported by the Jet Propulsion Laboratory - California Institute of Technology, under a contract with the National Aeronautics and Space Administration (80NM0018D0004). This work was partially funded by the Defense Advanced Research Projects Agency (DARPA). \textcopyright  2022 All rights reserved.}
\ral{\thanks{Digital Object Identifier (DOI): see top of this page.}}
}

%


\maketitle

\begin{abstract}
Multi-robot SLAM systems in GPS-denied environments require loop closures to maintain a drift-free centralized map.
With an increasing number of robots and size of the environment,
checking and computing the transformation for all the loop closure candidates becomes computationally infeasible. 
In this work, we describe a loop closure module that is able to prioritize
which loop closures to compute based on the underlying pose graph, the proximity to known beacons, 
and the characteristics of the point clouds.
We validate this system in the context of the DARPA Subterranean Challenge and \revfinal{on four challenging underground datasets  
where we demonstrate the ability of this system} to generate and maintain a map with low error.
We find that our proposed techniques are able to select 
effective loop closures which results in
51\% mean reduction in median error when compared to an odometric solution 
and 75\% mean reduction in median error when compared to a baseline version of this system with no prioritization.
\rev{We also find our proposed system is able to achieve a lower error in the mission time of one hour when compared to a system that processes every possible loop closure in four and a half hours.}
\arxiv{The code and dataset for this work can be found at \url{https://github.com/NeBula-Autonomy/LAMP}.}

\end{abstract}
\arxivheader{\thispagestyle{firstpage}}

%
\IEEEpeerreviewmaketitle

\section{Introduction}
\IEEEPARstart{S}{imultaneous} Localization and Mapping (SLAM) has become an essential 
piece in the perception stack of most robots and autonomous vehicles.
SLAM systems typically rely on two sources of information: odometry measurements (which describe the motion of the robot) and loop closures (which occur when a robot revisits an already seen location).

Loop closures are especially relevant with long operation times or large environments, 
since they are necessary in order to correct odometry drift 
when a robot revisits a previously seen part of the map.
In the multi-robot scenario, loop closures also serve the purpose of aligning the 
maps from individual robots
to create a consistent, drift-free map.
\revfinal{This is especially important in the context of the DARPA Subterranean Challenge where robots must find hidden artifacts in a large underground environment~\cite{darpa_subt}.}
\begin{figure}[t]
\centering
\includegraphics[trim=20 20 20 20, clip, width=0.9\columnwidth]{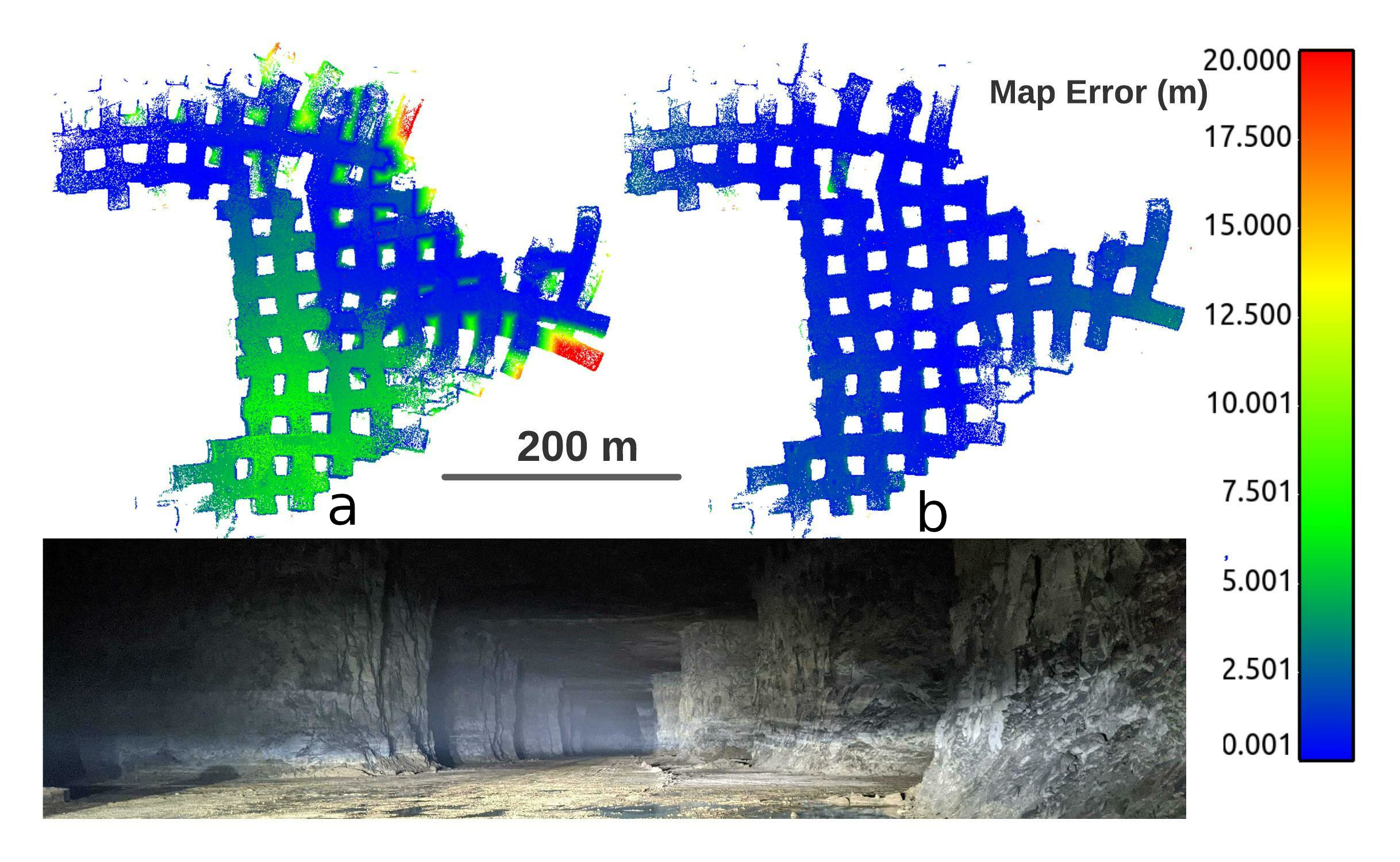}
\caption{
Demonstration of the benefit of our loop closure prioritization system, showing the accuracy of SLAM maps generated with (b) and without (a) loop closure prioritization. Data is from four robots driving over six km in the Kentucky Underground Mine.
We are only able to compute 0.5\% of the candidate loop closures. 
Picture taken within mine to demonstrate composition. 
\vspace{-5mm}}
\label{fig:lamp-ku-comp}
\end{figure}

\rev{For a multi-robot system operating in a large-scale environment, it is typically impractical to process every detected loop closure in the allotted mission time,}
since the number of possible candidates grows rapidly with time and the number of robots.
\rev{For example, in the Tunnel dataset (see \Cref{sec:expr_setup}) we have around 150,000 candidate loop closures.
As our system can process these at around 0.13 seconds per loop closure, it would take up to four and a half hours to process every loop closure, compared to a mission time of one hour.}
\rev{Additionally, not all loop closures provide improvement in the overall SLAM solution; in fact, many loop closures, such as those in parallel hallways, are false positives, and can cause the overall SLAM solution to degrade if these loop closures are added.}
To address this challenge, 
we propose a loop closure system that 
prioritizes certain loop closure candidates as to ensure that loop closures that are likely 
to be effective are computed first.
The system consists of three modules that each address the problem of choosing loop closures in different ways.
We develop this system in a centralized multi-robot SLAM framework~\cite{Ebadi2020LAMPLA,Lamp2}.

The first module approaches the prioritization problem by inspecting the underlying pose graph and predicting which loop closures will result in the greatest reduction in \rev{pose} uncertainty. 
The second module prioritizes loop closures by assessing the observability of the associated sensor data. In this work we focus on the lidar, where more observable point clouds have less ambiguity in geometry.
Higher observability reduces the chances of false loop closures, and increases the chances of accurate alignment.
The third module uses artificial, deployed beacons to provide high-priority loop closures. 
Here, we focus on deployed radio beacons that transmit a unique ID to provide unambiguous place detection \revfinal{(like~\citet{funabiki2020range})}, but using the Received Signal Strength Indicator (RSSI) for approximate proximity detection.



\ph{Contributions}
Our key contribution is a system for the selection and prioritization of loop closures 
in a large multi-robot lidar based SLAM system that is 
scalable in the number of robots and the size of the environment. 
To achieve this we
\begin{itemize}
    \item Develop approaches for loop closure prioritization based on the predicted error reduction of adding a loop closure to our system, on the observability of the point clouds in a loop closure candidate, and on location proximity informed by artificial radio beacons.
    \item Perform experiments demonstrating the performance of the three approaches above with multiple robots in challenging large-scale underground environments, and show a 75\% mean reduction in median error compared to a baseline with no prioritization. We also demonstrate that our approach is largely invariant to the computational power of the centralized server, and \rev{demonstrate that our system can achieve a lower error in the allotted mission time of one hour compared to a system which processes all possible loop closures in four and a half hours.} 
\end{itemize}

\section{Background and Related Work}

\ph{Point Cloud Registration}
In lidar based SLAM systems, 
loop closures are usually computed by estimating the transform between two point clouds 
using point cloud registration. 
This is either based on iterative techniques to align dense point clouds in the case of Iterative Closest Point (ICP)~\cite{segal2009_gicp}, 
or on the detecting and matching of local geometric structures in the case of feature-based registration method~\cite{ebadi2021dareslam,Dub2017SegMatchSB}.
\revfinal{Importantly, point cloud registration is an expensive operation and cannot be performed for every possible candidate in the case of large-scale, multi-robot SLAM. 
In this work, we prioritize loop closures before performing registration to maximize the accuracy improvements for the computation expended.}

\revfinal{
\ph{Loop Closure Prioritization Systems}
The system we present in this paper performs loop closure prioritization to support large-scale, multi-robot SLAM. At the time of publication we know of no other directly comparable system, however there is existing work in loop closure \textit{selection}~\cite{yang_graduated_2020,mangelson_pairwise_2018,app12115291}. These systems perform an outlier rejection function by selecting the most consistent inliers from a set of loop closures. To do this selection, however, these systems require the loop closure transforms to be computed first, a computationally expensive step. In contrast, our work selects and prioritizes loop closures before computation, to increase robustness and accuracy in a computationally efficient way.}

\ph{Active Loop Closure}
Another similar area to our system is active SLAM, where an agent will be guided to create a loop closure that helps the overall SLAM solution~\cite{carlone_active_2014,cadenapastpresent}. 
Our work is similar to these works methodologically because we estimate the utility of candidate loop closures but differs in that we are passively observing candidate loop closures. \revfinal{We also differ in our goal, active SLAM typically guides robot navigation, whereas our approach looks to maximize the efficiency of computation.} 

\ph{Pose Graph Optimization}
\revfinal{When assessing the impact of a loop closure, active SLAM approaches typically use a graph structure in the context of pose-graph SLAM. The graph nodes are robot poses, to be estimated, and the graph factors connect these nodes and represent measurements such as odometric information or loop closures~\cite{dellaert2012factor}. The combination of factors and poses leads to a commonly used optimization problem called Pose-graph optimization~\cite{cadenapastpresent}. While regularly used to great effect, pose-graph optimization can be costly to perform at a high rate, as is needed when evaluating the impact of many loop closure candidates.}

\ph{Graph Neural Networks} 
\revfinal{One approach to reduce the computational burden of pose-graph optimization is to compute an approximation. We utilize graph neural networks (GNN)~\cite{zhou_graph_2019} to rapidly perform approximate inference on pose graphs, allowing us to evaluate many loop closure candidates and select the best set.
In particular, we use a Gaussian Mixture model graph convolution, which generalizes convolutional neural networks to graph based domains using mixture models~\cite{monti_geometric_2016}.
This model learns an embedding of nodes in a graph by passing the node embedding as messages and integrating the edge features. }

\ph{Graph Neural Networks and SLAM}
\revfinal{Graph neural networks have been used in SLAM, factor graphs and other probabilistic models before, in a similar way to what we propose~\cite{yoon_inference_2019,talak2021neural,satorras_neural_2020}. The closest to our work is the application of GNNs for selecting actions which increase the localization certainty in target coverage~\cite{zhou_graph_2021}. 
Our work differs by selecting loop closure candidates to perform point cloud registration on, rather than selecting actions, as in~\cite{zhou_graph_2021}.}

\begin{figure}[t]
    \centering
    \includegraphics[width=.8\columnwidth]{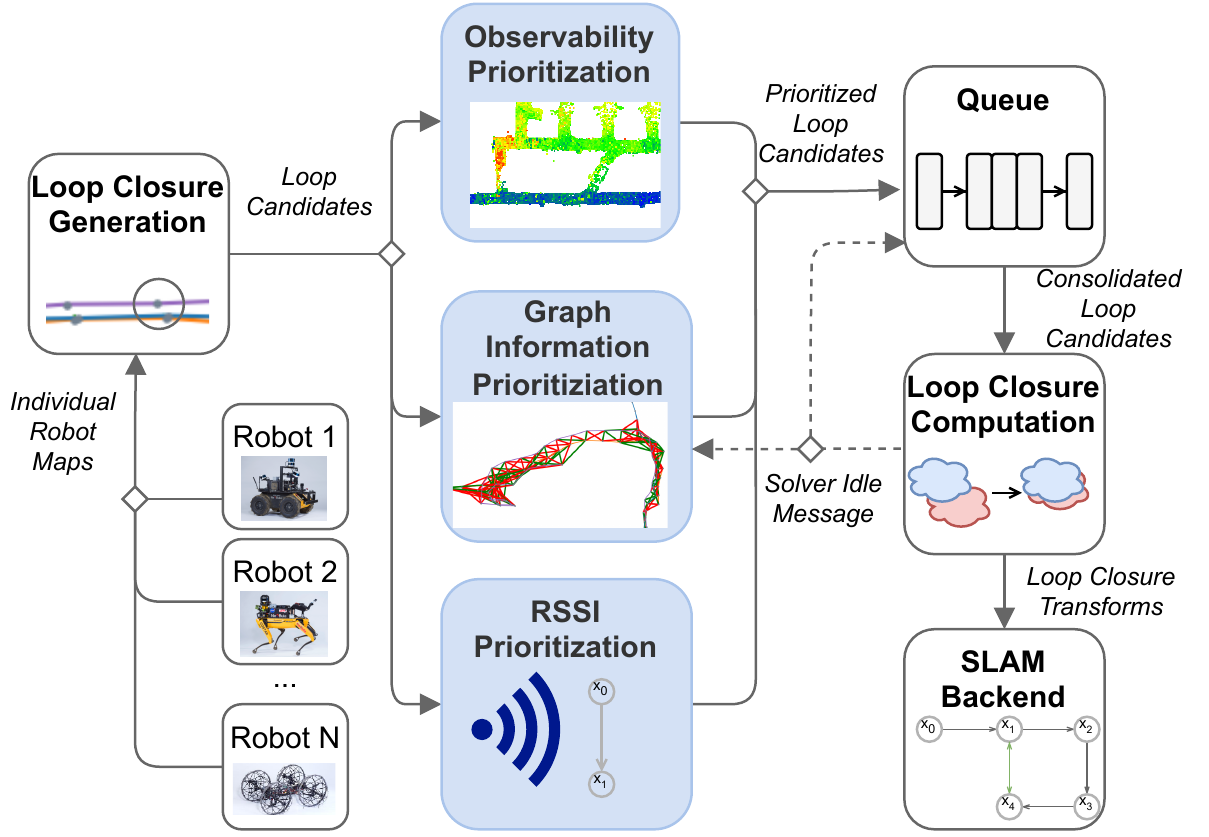}
    \caption{Diagram of SLAM multi-robot front-end and loop closure processing pipeline. Modules in blue are proposed in this work. Robot maps are merged on a base station and passed to the generation module, which performs proximity based analysis to generate candidate loop closures. These candidate loop closures are sorted by the prioritization modules, which feed into a queue. The loop closure solver computes the transformations between pairs of lidar scans which is given to the SLAM back-end which adds a loop closure factor. \vspace{-7mm}}\label{fig:lcd_diagram}
\end{figure}

\section{Approach}

In this section, we first contextualize the multi-robot SLAM front-end in LAMP~\cite{Ebadi2020LAMPLA,Lamp2},
the multi-robot SLAM system that our approach was tested in.
Then, we provide a description of the various steps and modules in the multi-robot SLAM front-end.
Finally, we highlight the prioritization modules, which are the main contribution of this work.

\subsection{SLAM System Overview}
Our multi-robot SLAM system, named LAMP, consists of three main components: the single-robot front-end, the multi-robot front-end,  and the multi-robot back-end.
The single-robot front-end runs on all the robots
and processes raw input sensor data to send pose graphs and keyed-scans to the base station.
The nodes in the pose graph represent estimated poses 
placed every $2 m$ traversal and the edges 
represent the relative odometric poses between between nodes. 
A keyed lidar scan is attached to each node in the pose graph and 
is the point clouds collected at the corresponding time.
The multi-robot front-end runs at the base station and consists of the loop closure detection system,
which collects the graph nodes and keyed-scans to produce loop closures. 
The multi-robot back-end then collects the graph produced by the single-robot front-ends and the loop closures
into a pose graph.
The back-end solves a pose graph optimization problem using
Graduated-Non-Convexity (GNC)~\cite{yang_graduated_2020}
for additional robustness to erroneous loop closures. 
We then stitch the keyed-scans together 
using the optimized poses to create the final LAMP map.
For more detail \revfinal{see~\citet{Ebadi2020LAMPLA,Lamp2}.}
\rev{The system does not assume that the robots are always in communication with the multi-robot base station. If a robot loses connection with the base station, it accumulates incremental pose graphs and keyed scans until communication has been regained.}
\rev{The system is initialized by aligning all robots to a common fixed frame using a specially designed calibration gate.} The full multi-robot SLAM front-end is shown in Fig.~\ref{fig:lcd_diagram}, 
and consists of three main steps: loop closure generation, loop closure prioritization, 
and loop closure computation.
\rev{All the SLAM front-end modules run in parallel and operate in a queued message passing system.}

\rev{\ph{Loop Closure Generation}}
While there are numerous ways to generate potential loop closures, such as place recognition~\cite{Cattaneo2021LCDNetDL,Schaupp2019OREOSOR,Cop2018DelightAE} or junction detection~\cite{ebadi2021dareslam}, we use proximity based generation in our experiments. This approach generates candidates from nodes that lie withing a certain distance $d$ from the most recent node
in the pose graph; $d$ is adaptive and is defined as 
$d = \alpha|n_{current} - n_{candidate}|$, which is dependent on the relative traversal between two nodes for the single-robot case and 
$d = \alpha n_{current}$, which is dependent on the absolute traversal for the multi-robot case, 
where $n_{current}$ and $n_{candidate}$ are the index of the current and candidate nodes respectively and $\alpha$ is a constant ($0.2m$).
\rev{Loop closures are generated when new keyed scans are sent from a robot to the base station. If the robots are not in communication range, the prioritization and computation nodes continue to work on the backlog of loop closures.
}

\rev{\ph{Loop Closure Prioritization} In order to select the best loop closures to process during the mission execution we propose three independent modules that orders the loop closures to process in batches.}
The \textbf{Graph Information Prioritization} module, which we will discuss in Section~\ref{sec:batcher}, uses a graph neural network to determine which loop closure candidates would decrease the error the most if they were added to the graph.
The \textbf{Observability Prioritization} module, which we will discuss in Section~\ref{sec:observability}, prioritizes loop closure candidates with point clouds that
have many geometric features
and are particularly suitable for ICP-based alignment.
The \textbf{RSSI Prioritization} module, which we will discuss in Section~\ref{sec:rssi}, attempts to find loop closures which are close to a known radio beacon.

A {\bf queue} is used to combine the loop closure candidates suggested from the prioritization modules, 
\rev{using} a simple round-robin queue which takes loop closure candidates from each prioritization module equally. Once these loop closure candidates are ordered, they are stored until the loop closure computation module is done computing the transforms of the previous loop closures. 
\rev{Each time the loop closure computation node is done processing the previous set, the queue sends another fixed-size set of loop closures for computation. 
This allows the generation and computation rates to be independent, and the loop closure computation module to be saturated as often as possible.}

\rev{\ph{Loop Closure Computation}} performs a two-stage process to compute the loop closure transform:
first we use SAmple Consensus Initial Alignment (SAC-IA)~\cite{Rusu2009FPFH}, 
a feature-based registration method, 
to first find an initial transformation,
and then perform Generalized Iterative Closest Points (GICP)~\cite{segal2009_gicp} to refine the transformation.
We discard the loop closures that either have a mean error that exceeds some maximum threshold after SAC-IA ($32 m$ in our experiments) 
or have a mean error that exceeds some maximum threshold after GICP ($0.9 m$ in our experiments).
The remaining loop closures are then sent to the back-end,
and the computation module requests more loop closures to 
both the queuing and the Graph Information Prioritization modules.
To compute the loop closures in an efficient and scalable manner, we use an 
adaptable-size thread pool to perform computation in parallel across the current consolidated loop closure candidates.
The most computationally expensive step in this system is the loop closure computation step.
The prioritization approach is designed to achieve maximum benefit from computation expended in loop computation, by only processing the loop closure candidates most likely to succeed (\textbf{Observability}, \textbf{RSSI}), and improve localization (\textbf{Graph Information})

\subsection{Graph Information Prioritization}\label{sec:batcher}
\rev{The graph information prioritization module seeks to find sets of loop closures that reduce the uncertainty in the poses in the underlying pose graph. 
The graph information module selects a set of $B$ loop closures each time the loop closure computation module is idle, selecting the next set of $B$ loop closures while the loop closure computation module processes the current set.
The graph information prioritization module continues generating sets of loop closures until the mission ends.}
We propose a system which analyzes the desirability of a loop closure by learning to estimate the reduction in uncertainty attained by adding a set of loop closures. 
Formally, we seek to minimize
    $f(l) =  \argmin_{l \subset L} \gnnobj s.t. |l| = B$
where $PG$ is a pose graph, $L$ is the set of all the current candidate loop closures, $B$ is some fixed amount of loop closures per batch, and $e$ is the objective function which is the trace of the covariance of the pose estimates after pose graph optimization.
Directly optimizing this equation is difficult for two reasons. 
The first difficulty is that computing $\gnnobj$ directly requires solving the pose graph with each hypothetical set of loop closures. 
This is a large computational cost for each evaluation of the objective function.
Secondly, there are numerous possible combinations of $l \subset L$, more precisely $O(L^B)$.  

\ph{Computing the Objective Function}
\rev{Computing $\gnnobj$ requires solving the underlying pose graph, which is prohibitively expensive for optimization as this objective must be queried many times.}
To solve the problem of computing $\gnnobj$ efficiently, we train a graph neural network (GNN) which learns to predict the covariance trace after the pose graph is solved. 
We call this estimate of the objective $\gnnobjest$.
\rev{The graph neural network is quicker to evaluate than solving the underlying pose graph, allowing us to compute the objective function online many times for optimization.
We found that in the Tunnel dataset (see \Cref{sec:expr_setup}) the graph neural network takes on average 0.8 seconds to evaluate while the pose graph optimizer takes an average of 15 seconds.}

The input to the graph neural network is constructed by adding graph neural network nodes for each pose in the pose graph.
The features of these nodes are the translation of the pose as well as the translational covariance for each pose, \revfinal{in the world frame}.
The graph neural network edges are constructed by taking the factors between poses and labeling each edge with a one hot encoding of its factor type, such as loop closure or odometry.

The graph neural network is constructed by three layers of graph convolutions followed by a global additive layer and four dense 64 neuron wide layers interleaved with batch normalization layers. 
In this work, we use the Gaussian Mixture model layer with four kernels for the GNN layers~\cite{monti_geometric_2016}.
This layer type is chosen because of its ability to handle edge features, such as the type of factor.

To train the graph neural network we compile a dataset of pose graphs solved with various loop closures from previous simulated and real datasets.
The target output of the graph is the trace of the covariances of the pose graph after it is solved by the back-end, equivalent to $\gnnobj$.
By using a graph neural network, we are able to compute $\gnnobj$ much more efficiently, albeit with less accuracy.

\ph{Minimizing the Objective Function}
We aim to minimize $\gnnobj$ by selecting a set of loop closures $l$. 
\rev{This set of $|l|=B$ loop closures is chosen while the loop closure computation module aligns the previous set of loop closures.
The number of loop closures in each set, $B$, is chosen to trade off between the time it takes to optimize the set of loop closures and the delay in incorporating new candidate loop closures into $L$, the set of all current candidate loop closures being optimized over.}
\rev{To minimize the objective function we modify the simulated annealing optimization algorithm~\cite{simulated_annealing}.
This algorithm maintains a set of loop closures and iteratively chooses new sets of loop closures which decrease the estimated error. 
\revfinal{Simulated Annealing is used to explore sets of loop closures stochastically without the use of gradient information. 
The cooling constant, $c$, determines the probability of accepting a worse solution than the current one. 
This acceptance probability exponentially decreases with the number of steps.}
Simulated annealing is an anytime algorithm, which is required as we do not know ahead of time how long the current set of loop closures will take to process in the loop closure computation module, and would like to improve our solution in parallel while the previous set of loop closures are processed.}
Simulated annealing has been shown to be effective at minimizing information theoretic objectives, such as in sensor positioning~\cite{leitold_network_2018,saoptimalsensor}.
We modify the standard simulated annealing algorithm in a few ways and our adapted algorithm can be seen in \Cref{alg:batcher}.
\begin{figure}[t]
    \centering
    \includegraphics[ trim={1px 1px 1px 1px}, clip, width=.6\columnwidth,]{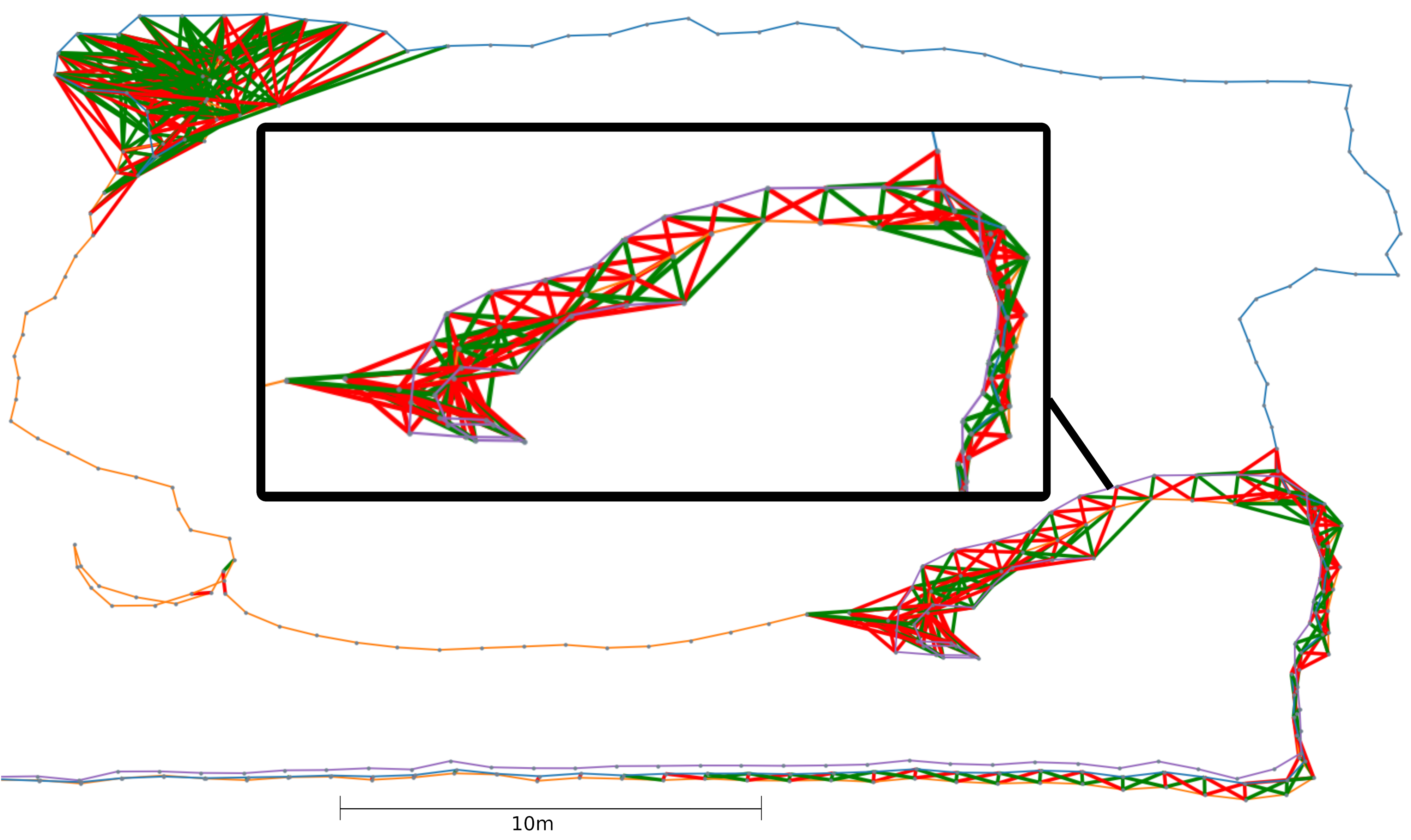}
    \caption{Example output of the Graph Information Prioritization module on the \textbf{Urban} dataset. Green lines are selected loop closures, red are not selected. Orange, blue and purple edges are individual robot odometry. }\label{fig:batcher}
    \vspace{-0.1in}
\end{figure}
\begin{algorithm}[t]
\caption{Graph Information Prioritization Algorithm \\
\scriptsize{$PG$ is the current pose graph, $L$ is the set of candidate loop closures, $T_{max}$ is the starting temperature, $T_{min}$ is the final minimum temperature, $c$ is the cooling constant, $step$ is the current iteration, $U$ is the uniform distribution} }\label{alg:batcher}
\begin{algorithmic}[1]
\State $l\gets$ HEURISTIC$(PG,L)$ \Comment{Initial state}
\State $err \gets \tilde{e}(PG \cup l)$ \Comment{Initial error}
\While{loop closure computation module is not idle} 
    \State $T \gets max(T_{max} c^{(step)}, T_{min})$ \Comment{Temperature update}
    \State $l' \gets$ EDGE\_SWAP$(l,L \setminus l)$ \Comment{Select new state}
    \State $err' \gets \tilde{e}(PG \cup l')$ \Comment{Compute new error}
    \State $\delta = err' - err$  \Comment{Error change of new state}
    \If{$\delta  < 0.0$ \textbf{or} $e^{-\frac{\delta }{T}} > U[0,1]$} 
        \State{$l, err \gets l', err'$} \Comment{Accept new state}
    \EndIf 
\EndWhile
\State $l \gets$ BEST$(l)$ \Comment{Return state with lowest error}
\end{algorithmic}
\end{algorithm}
First, we use a HEURISTIC function to select the initial \rev{set of loop closures}.
We compared a few different heuristics and selected one which attempts to maximize the spatial distance between the loop closures.
Secondly, we weigh the probability of swapping a pair of loop closures between the current solution and the set of all loop closures by the inverse of their distance, in the EDGE\_SWAP method.
This penalizes swapping a loop closure for another loop closure which is a large distance away.
This encourages each set of loop closure produced to be spatially diverse. 
This spatially distanced heuristic is designed to avoid having sets of loop closures only focused on a single high uncertainty junction, and instead add loop closures at different locations to reduce the overall graph uncertainty.
When the loop closures are aligned by the solver, a message is generated which stops the optimization process and produces a solution\rev{.
The Graph Information Prioritization module then generates another set of loop closures in parallel to the loop closure computation module aligning the current set.}

We show an example output of the Graph Information Prioritization module in \cref{fig:batcher}. 
Loop closures are spatially spread out where there is overlap of trajectories. 
In this example, the Graph Information Prioritization module selects $B=250$ loop closures from a set of $|L|=500$ loop closures.

\subsection{Observability Prioritization}\label{sec:observability}
Perceptual aliasing is the problem of mistaking two non-overlapping locations as overlapping in a lidar scan due to the presence of similar perceptual features.
This is especially common in long feature-less corridors.
Filtering out the feature-poor loop closure candidates 
saves us from spending computationally expensive point cloud registration time 
on error-prone loop closures. 
Prioritizing feature-rich areas allows us to perform point cloud registration 
on areas that will more likely return better loop closures.
To quantify how feature-rich or feature-poor a point cloud scan is, 
we draw inspiration from the \textit{observability} metric and \textit{inverse condition number} presented in~\cite{Tagliabue2020LIONLO, Zhang2016Degeneracy} where 
the observability is described in terms of the minimum eigenvalue of the Information matrix
$\mathbf{A}$ of the Point-To-Plane ICP cost for two clouds $i$ and $j$. 
\begin{equation}\label{eq:inf_mtx}
    \mathbf{A} = \sum_{i=0}^{n} \mathbf{H}_k^{\top}\mathbf{H}_k
\end{equation}
In \cref{eq:inf_mtx}, $n$ is the number of correspondences and $\mathbf{H_k}$ is the residual Jacobian, which can be derived as, 
\begin{equation}\label{eq:jacob_mtx}
    \mathbf{H}_k = \begin{bmatrix}(\mathbf{p}_j^{(k)} \times (\mathbf{R}^{*}_{ij})^\top \mathbf{n}^{(k)}_i)^\top && ((\mathbf{R}^{*}_{ij})^\top \mathbf{n}_i^{(k)})^\top \end{bmatrix}
\end{equation}
where $\mathbf{p}_j^{(k)}$ is the position of the $k$-th point correspondence in cloud $j$, 
$\mathbf{R}^{*}_{ij}$ is the rotation computed by ICP, 
and $\mathbf{n}^{(k)}_i$ is the normal of the $k$-th correspondence in cloud $i$.

\begin{figure}  
\centering
\includegraphics[width=.6\columnwidth, trim= 8mm 10mm 6mm 10mm, clip]{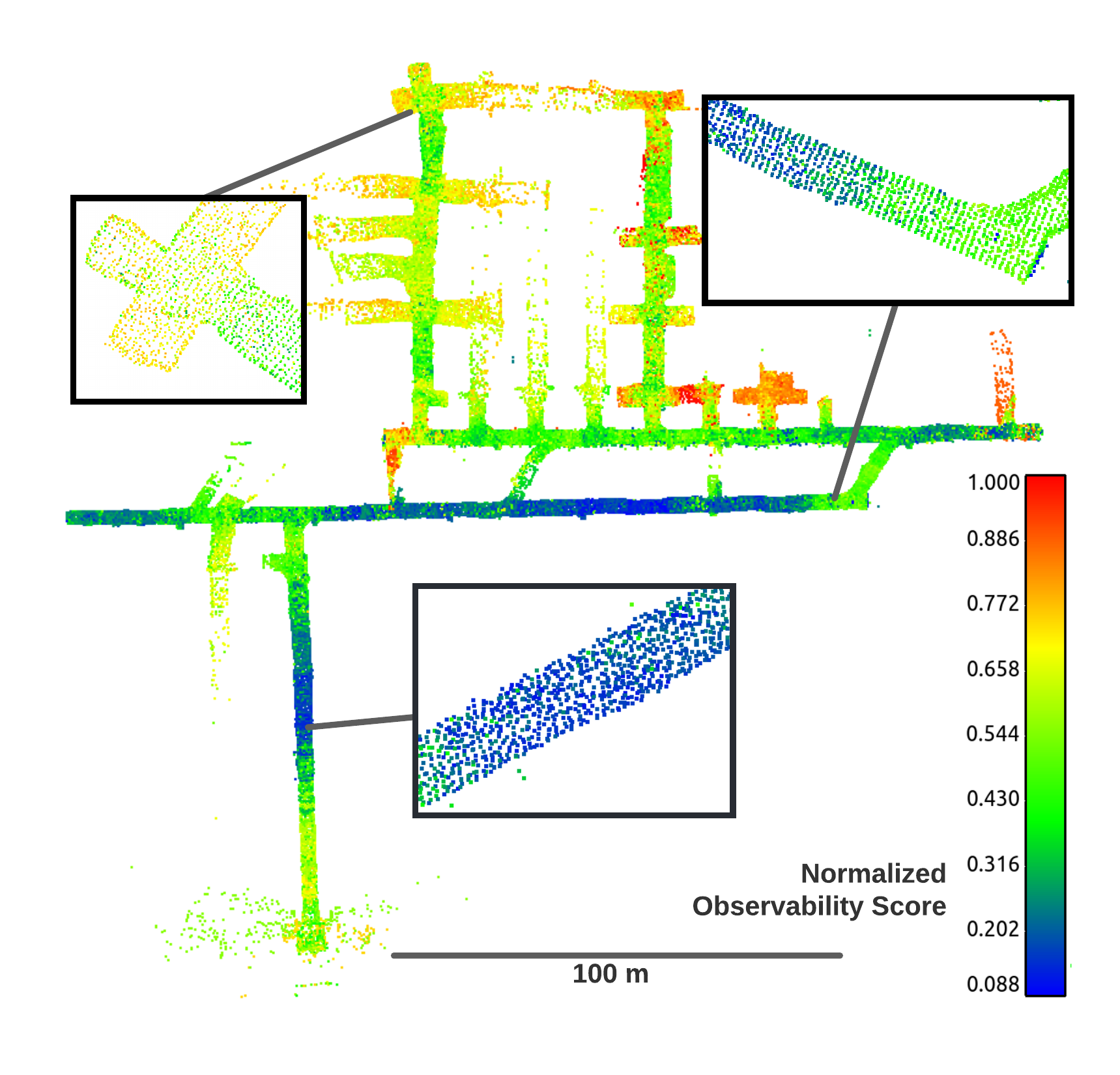}
\caption{Visual illustration of point cloud map with intensity colored by the normalized observability score. Note how the lower observability areas are usually those of long straight corridors with less features. With observability prioritization, we prioritize loop closures at corners and junctions: 
which are less susceptible to perceptual aliasing. \vspace{-4mm} } 
\label{fig:observability_demo}
\end{figure}

Since we want to quantify the observability before executing ICP the rotation $\mathbf{R}^{*}_{ij}$ in \cref{eq:jacob_mtx} is unknown. 
However, we observe that for two identical point clouds, we can assume that $\mathbf{R}^{*}_{ij} = I_3$.
Hence, we define \textbf{observability for a single point cloud} based on the eigenvalues of the matrix $\tilde{\mathbf{A}}$,
\begin{equation}
    \tilde{\mathbf{A}} = \sum_{i=0}^{n} \tilde{\mathbf{H}}_k^{\top}\tilde{\mathbf{H}}_k
\end{equation}
where $n$ is the number of points in the point cloud and 
\begin{equation}
    \tilde{\mathbf{H}}_k = \begin{bmatrix}(\mathbf{p}^{(k)} \times \mathbf{n}^{(k)})^\top && (\mathbf{n}^{(k)})^\top \end{bmatrix}
\end{equation}
We define the \textbf{observability score} of a scan cloud 
as the minimum eigenvalue of the observability matrix $\tilde{\mathbf{A}}$ 
and the \textbf{normalized observability score} as the observability score 
divided by the largest observability score calculated up to the current point in time 
(i.e. the first point cloud will always have a normalized observability score of 1). 
Fig.~\ref{fig:observability_demo} shows an example map colored by the normalized observability score. 
We observe that long corridors have low observability scores, while intersections and feature-rich areas have high observability scores.
For prioritization, we first filter out the candidates for which the sum of the 
normalized observability scores of the two point clouds involved in the loop closure 
are below a certain threshold. 
The remaining candidates are then sorted based on the normalized observability score sum, 
and the candidates with the higher scores are prioritized in adding to the candidate queue 
for loop closure computation.

\subsection{RSSI Prioritization}\label{sec:rssi}
\begin{figure}
	\centering     
	\includegraphics[trim={0cm 7.2cm 0cm 1cm},clip, width=0.5\columnwidth]{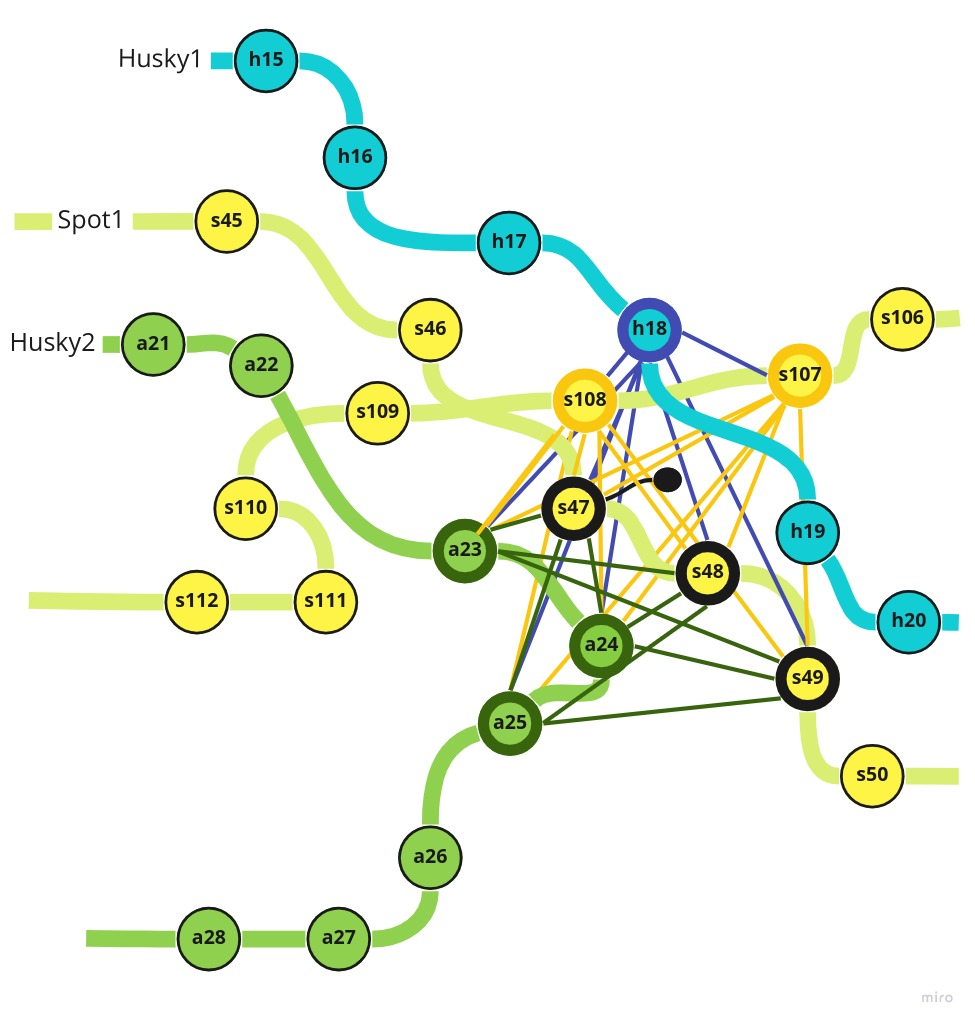}
	\caption{
        An example RSSI-Based loop closure scenario. Robot Spot1 deploys the communication node (black). Other robots (Husky1 and Husky2) perform a flyby of the RSSI node, and Spot1 returns for a second flyby, generating many known loop closures, shown as the thin lines.
		\vspace{-5mm}}
	\label{fig:rssi_loop}
	\vspace{-.25cm}
\end{figure}

Radio beacons carried by robots and dropped during unknown environment exploration allow communication between robots when not in range of the base station~\cite{agha2021nebula}.
The beacons act not only as a mesh of communication medium, but also as an indicator that a robot is near a location where a beacon has been deployed (each radio has a unique ID).
Once a beacon is deployed, all graph nodes within a threshold RSSI signal strength are added as one set loop candidate nodes. When a robot deploys the beacon, there are no loop closures, but for every subsequent pass of a robot (called a "fly-by"), there are two, or more, sets of loop candidate nodes to match as loop candidates.
\rev{This process can be seen in \cref{fig:rssi_loop} which shows the loop closures generated for multiple robots passing a single beacon.}
\rev{We process these fly-by nodes in the order that the robot traverses them.}

The RSSI module uses the path signal loss concept for pseudo-range measurements to approximate distance between a transmitter (deployed in an environment) and a receiver (carried by a robot).
Path loss is the reduction in the power density of an electromagnetic wave as it propagates through space.
Path loss between a robot and a beacon is defined as the difference between the transmitted power $T_{actual~dBm_{k}}$ and received signal power $T_{received~dBm_{k}}$ for each receiver $k$, \rev{averaged over each receiver.}
The loop closure is added to the flyby when the estimated distance is less than $threshold$:
\begin{equation}
    \frac{1}{N} \sum_{k=1}^{N}
	 (T_{actual~dBm_{k}} - T_{received~dBm_{k}})
	 < threshold
\end{equation}

\section{Experiments}
We demonstrate the performance of loop closure prioritization in challenging environments.
We show that our system reduces the error when compared to a baseline system.
This baseline system is the older version of LAMP, presented in ~\cite{Ebadi2020LAMPLA},
which does not include any loop closure prioritization and process loop closure candidates for loop closure computation \rev{in the order they are generated}.
We also compare against a solution which only uses the odometric data from the robots and does not perform any loop closures.
We perform an ablation study of our system to demonstrate combined power of our proposed modules.

\subsection{Experimental Setup}\label{sec:expr_setup}

We evaluate the loop closure prioritization modules on four datasets collected by Team CoSTAR.
The first dataset is the \textbf{Tunnel Dataset} 
and includes two Huskies traversing up to 2.5 km combined, in a coal mine that consists of mostly featureless narrow tunnels.
The second dataset is the \textbf{Urban Dataset} 
and includes two Huskies and a Spot traversing up to 1.5 km combined in an abandoned nuclear power plant 
consisting of a two-floor environment with open areas, small rooms, narrow passageways, and stairs.
The third dataset is the \textbf{Finals Dataset}
and includes three Spots
and a Husky traversing up to 1.2 km combined in the DARPA-built course including tunnel, cave, and urban-like environments.
The last dataset is the \textbf{Kentucky Underground Dataset}
and includes four Huskies traversing up to 6 km combined
in a limestone mine, which consists of large 10-20 m wide tunnels.
\revfinal{We are only able to evaluate the RSSI prioritization module on the \textbf{Finals Dataset} due to the requirement of deploying radio beacons.}

\rev{
We compare a variety of systems in both the ablation study (\Cref{sec:ablation}) and the system study (\Cref{sec:system_results}).
\textit{Bsln} is the baseline system which does not have any loop closure prioritization and processes the loop closures in the order they arrive. 
\textit{Obs} only uses Observability prioritization, \textit{GI} uses only the Graph Information priorizitation and \textit{Full} uses both prioritization methods.
\textit{Rand} selects loop closures randomly.
\textit{Odom} does not use any loop closures and only uses the odometry measurements from the robots.
}

The experiments are performed using the recorded field data
on a laptop with an Intel i7-8750H processor with 12 cores.
To compare results on a machine with higher computational power, 
we also perform a set of experiments on a powerful server, 
which has an AMD Ryzen Threadripper 3990x processor with 64 cores.

\subsection{Ablation Results}\label{sec:ablation}
We perform an ablation study by running LAMP~\cite{Ebadi2020LAMPLA} 
on the centralized server 
with no loop closure prioritization, with the individual prioritization modules, 
and with the complete system.
\rev{For each test, we allow the system to run only for the duration of the mission (i.e. we stop the run and record the results as soon as there are no more new sensor messages).}
The full ablation results are shown in Table ~\ref{tab:lc-ablation}, 
which shows in detail the amount of loop closures, as a 
percentage of the loop candidates that are added to the computation queue. \rev{Since prioritization also prunes a number of candidates, in general, the number of loop candidates in the computation queue is higher for the baseline system.}
The categories are: passing loop closure computation ("Verified"), passing GNC as inliers ("GNC-Inliers"), and having error under a threshold (0.05m, 0.05 rad) to the ground truth ("Inliers").
We also include the final absolute trajectory error (ATE) for context.

Notice that with prioritization, 
we are able to process significantly more loop closures that 
end up passing computation and are true inliers, with the exception of the Tunnel dataset, \rev{since many loop closure candidates are left unprocessed without prioritization.}
Experiments with observability prioritization show a higher percentage of loop closures that pass computation
because observability prioritization prioritizes loop closure candidates that are suitable for ICP-based alignment.
In all the datasets, with either graph information prioritization or observability prioritization, or both,
we are able to process more inliers;
the higher percentage of true inliers shows that we are able to process a larger number of accurate loop closure with prioritization, 
and the higher percentage of GNC inliers shows that more loop closures survive outlier rejection
to contribute to refining the trajectory estimate of the SLAM system,
which reflects in the better trajectory estimates for all of the datasets except for Urban when using prioritization.

\begin{figure}
\centering
\subfloat[Tunnel]{\includegraphics[trim=20 0 20 20, clip, width=0.49\columnwidth]{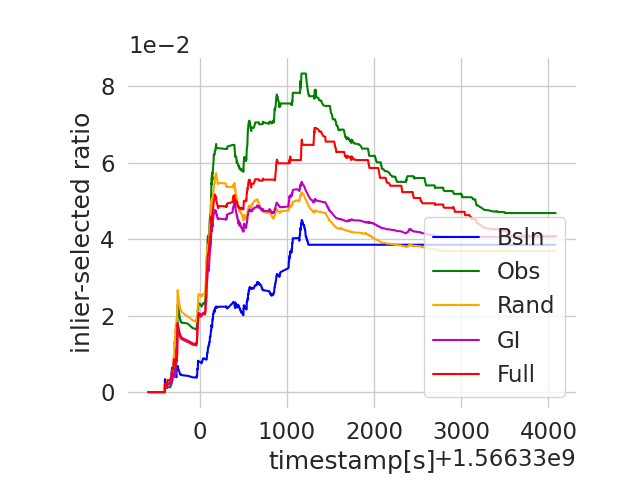}}
\hfill
\subfloat[Kentucky Underground]{\includegraphics[trim=20 0 20 20, clip, width=0.49\columnwidth]{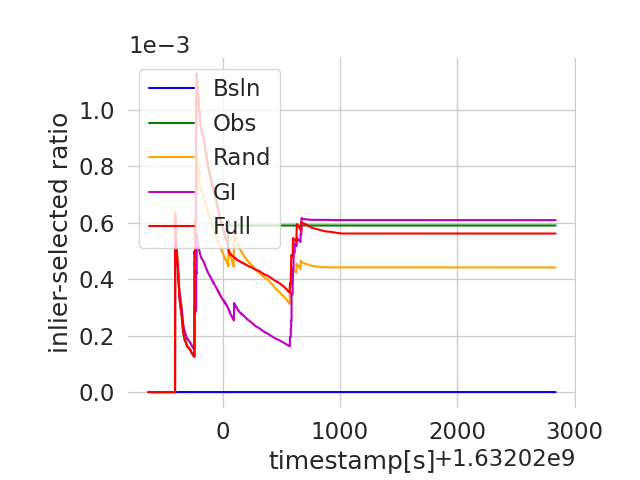}}
\caption{Selected loop closure candidate to inlier ratio with the different prioritization methods. With no prioritization ("Bsln"), 
little to no inliers are detected from the candidates in the latter part of the run for the Urban, Finals, and Kentucky Underground datasets. }
\label{fig:inlier-selected-ratio}
\vspace{-8mm}
\end{figure}

Fig.~\ref{fig:inlier-selected-ratio} shows the inlier-to-queued loop closure ratio over the course of the run. We see that for the baseline method without prioritization, the ratio is usually heavily biased towards the beginning, since the loop closures are computed in the order of generation, and the system will not compute any loop closures in the later part of the mission before the end of the mission. With loop closure prioritization, we are able to maintain a reasonable ratio for the entire length of the experiment, showing that we are able to successfully detect loop closure \emph{that are inliers} across the complete length of the experiment.
Moreover, the higher inlier-to-queued ratio for the experiments running the prioritization 
methods show that we are using less time on loop closure computation for the outliers, 
while the low ratio on the baseline means that most computation is spent on outliers.

\begin{figure}[t]
\centering
\subfloat[Loop closures \revfinal{accepted} over time]{\includegraphics[trim=20 0 30 35, clip, width=0.49\columnwidth]{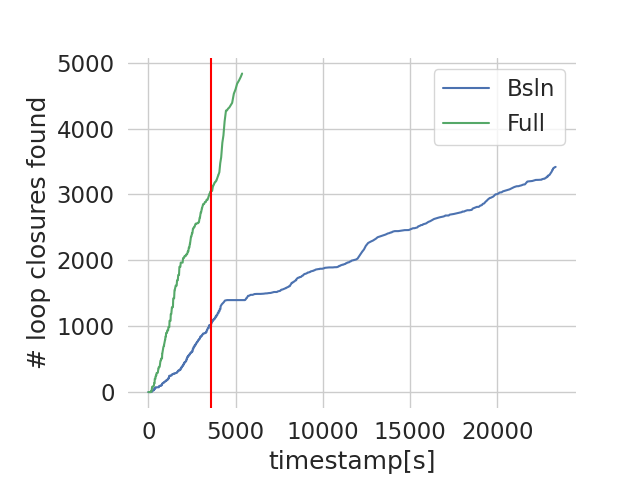}}
\hfill
\subfloat[Average trajectory error over time]{\includegraphics[trim=20 0 30 35, clip, width=0.49\columnwidth]{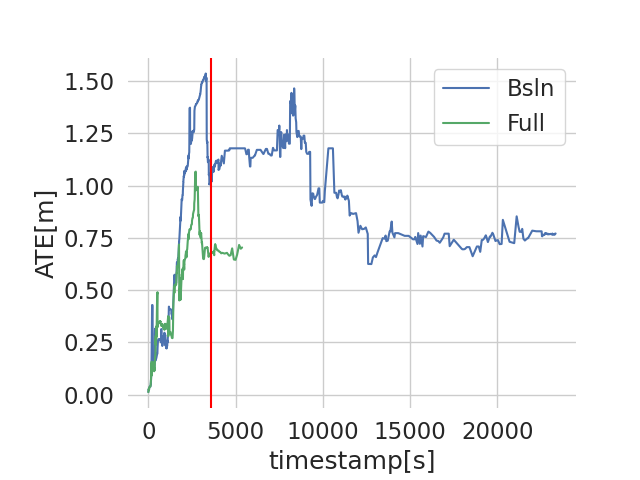}}
\caption{\rev{Direct comparison of with and without prioritization on the Tunnel dataset. The red line marks the end of the mission. The full system is able to find a similar number of inlier loop closures and a lower error at the end of the mission than the baseline can after computing all loop closures.} \vspace{-7mm}}
\label{fig:extended-exp}
\end{figure}

\rev{In \cref{fig:extended-exp} we further showcase the importance of prioritization. 
On the tunnel dataset, we allowed the system to run past the duration of the mission
and plot the number of loop closures detected along with the trajectory error. With prioritization, we were able to detect more loop closures in less than one-quarter of
the time compared to without prioritization. 
The trajectory error is also reduced earlier due to the earlier detection of the loop closures, giving us a better trajectory estimate before the conclusion of the mission.}

\setlength{\tabcolsep}{4pt}
\begin{table}[t!]
\centering
\caption{ATE and Verified or inlier loop closures for different prioritization methods as \% of queued loop closures.}\label{tab:lc-ablation}
\begin{tabular}{cl ccccc}
\toprule
& & Bsln & Obs & Rand & GI & Full \\
\midrule
\multirow{4}{*}{Tunnel}
& Verified (\%) & \textbf{18.192} & 13.816 & 14.353 & 14.406 & 13.445 \\
& GNC-Inliers (\%) & 3.724 & \textbf{7.973} & 5.838 & 6.300 & 7.200 \\
& Inliers (\%) & 3.851 & \textbf{4.679} & 3.684 & 4.056 & 4.082 \\
& ATE (m) & 1.12 & 0.75 & \textbf{0.72} & 0.77 & 0.94 \\
\midrule
\multirow{4}{*}{Urban}
& Verified (\%) & 3.578 & \textbf{9.200} & 6.421 & 6.017 & 6.319 \\
& GNC-Inliers (\%) & 0.518 & \textbf{4.312} & 2.460 & 2.271 & 2.213 \\
& Inliers (\%) & 0.06 & 0.899 & \textbf{1.098} & 0.858 & 0.928 \\
& ATE (m) & 0.93 & \textbf{0.93} & 0.99 & 1.00 & 1.06 \\
\midrule
\multirow{4}{*}{KU}
& Verified (\%) & 0.062 & 0.555 & 0.418 & 0.539 & \textbf{0.561} \\
& GNC-Inliers (\%) & 0.009 & \textbf{0.344} & 0.247 & 0.2293 & 0.334 \\
& Inliers (\%) & 0.0 & 0.059 & 0.044 & \textbf{0.061} & 0.056\\
& ATE (m) & 4.41 & 6.02 & 3.87 & 3.32 & \textbf{3.00} \\
\midrule
\multirow{4}{*}{Finals}
& Verified (\%) & 4.105 & \textbf{5.462} & 5.382 & 4.953 & 4.614 \\
& GNC-Inliers (\%) & 0.271 & 1.461 & 1.435 & 1.318 & \textbf{1.466} \\
& Inliers (\%) & 0.242 & 1.219 & \textbf{1.224} & 1.065 & 0.97 \\
& ATE (m) & 0.69 & 0.27 & 0.34 & 0.32 & \textbf{0.23} \\
\end{tabular}

\vspace{-5mm}
\end{table}

\subsection{System Results}\label{sec:system_results}
\begin{figure}
\centering
\subfloat[Tunnel]{\includegraphics[trim=20 0 30 20, clip, width=0.49\columnwidth]{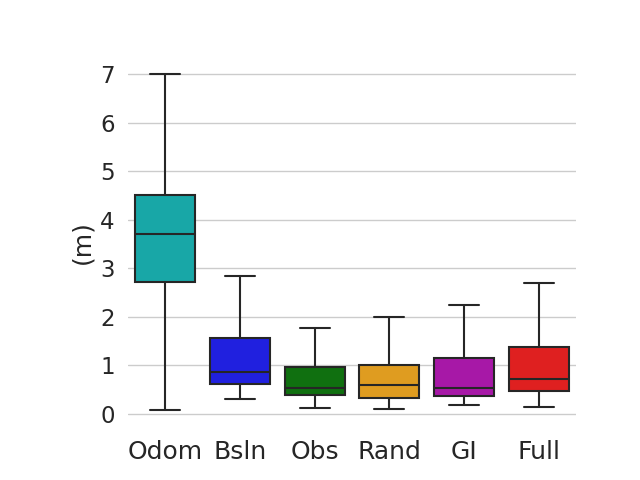}}
\hfill
\subfloat[Urban]{\includegraphics[trim=20 0 30 20, clip, width=0.49\columnwidth]{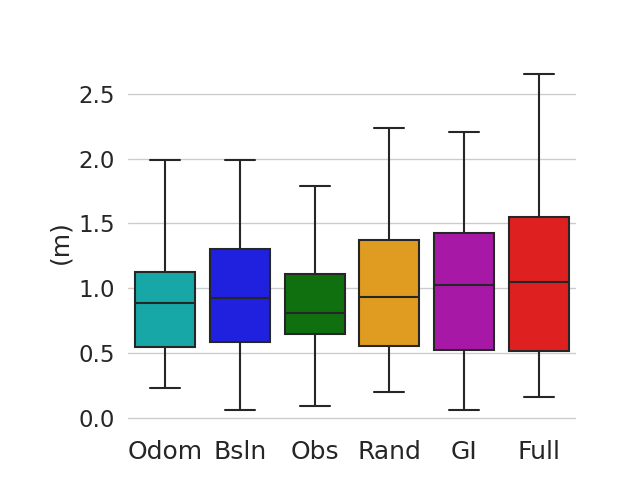}}
\hfill
\subfloat[Finals]{\includegraphics[trim=20 0 30 20, clip, width=0.49\columnwidth]{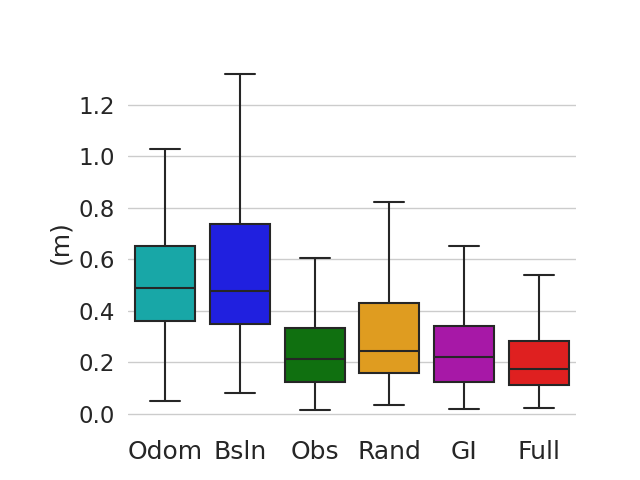}}
\hfill
\subfloat[Kentucky Underground]{\includegraphics[trim=20 0 30 20, clip, width=0.49\columnwidth]{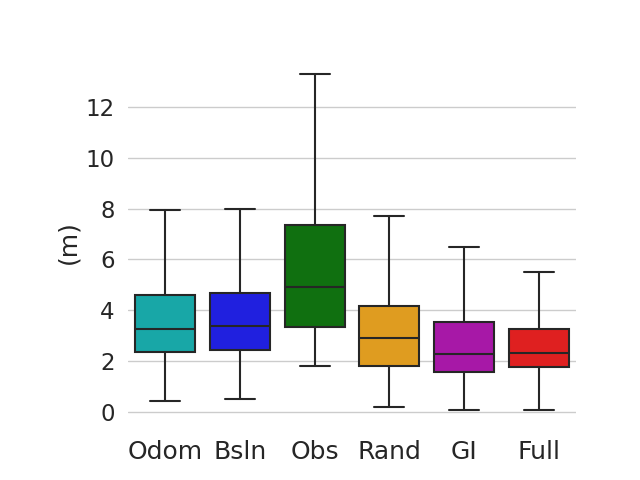}}
\caption{Trajectory error comparison with for consumer laptop  on the four underground datasets. \rev{We find the full system is able to consistenly outperform the baseline, odometry, and random solutions.} }
\label{fig:ate-boxplots}
\vspace{-7mm}
\end{figure}

The comparisons of the final trajectory error with the different configuration are shown in Fig.~\ref{fig:ate-boxplots}, 
showing that any prioritization method improves the final result relative to the baseline.
Fig.~\ref{fig:lamp-ku-comp} compares the final mapping results without loop closure prioritization
and with our full proposed system which provides a large increase in performance.
The results also show that the combination of observability and graph information prioritization improve overall system performance. For instance, in the Kentucky Underground environment, we find that observability performs more poorly than a policy which takes random loop candidates but that graph information performs well, causing the full system to perform well.
In all environments except Urban the Graph Information Prioritization module has better median performance than randomly selecting edges, the baseline and odometry-only solutions.
The same is true for Observability Prioritization in all environments except the Kentucky Underground.
\rev{We find that in the Finals dataset, the baseline performs worse than an odometry only solution. We believe this is because there are many false positive loop closures which make the overall SLAM solution worse.}

\begin{figure}
\centering
\subfloat[Tunnel]{\includegraphics[trim=20 0 30 20, clip, width=0.49\columnwidth]{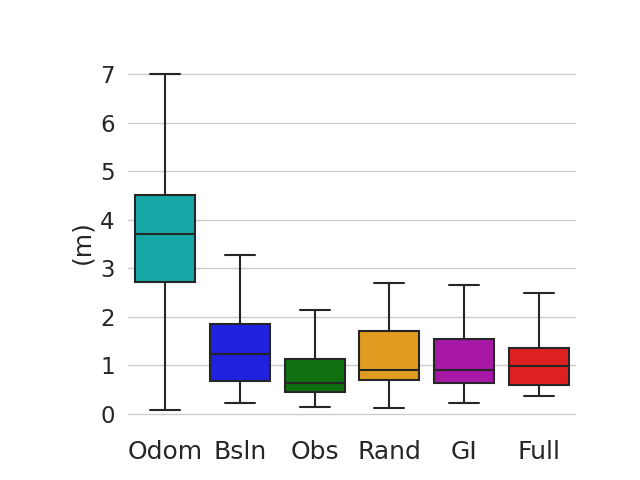}}
\hfill
\subfloat[Urban]{\includegraphics[trim=20 0 30 20, clip, width=0.49\columnwidth]{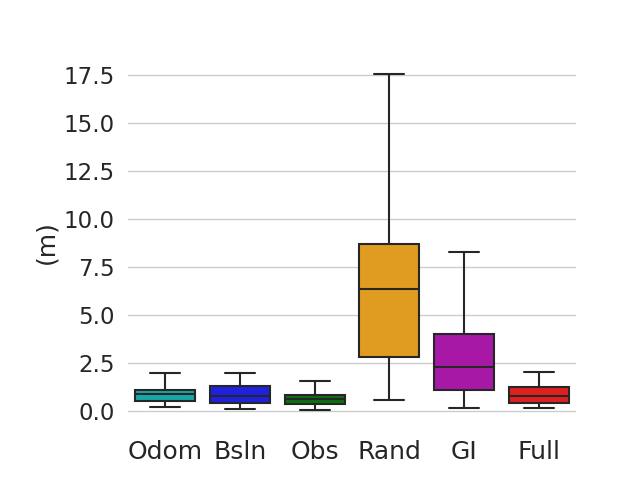}}
\hfill
\subfloat[Finals]{\includegraphics[trim=20 0 30 20, clip, width=0.49\columnwidth]{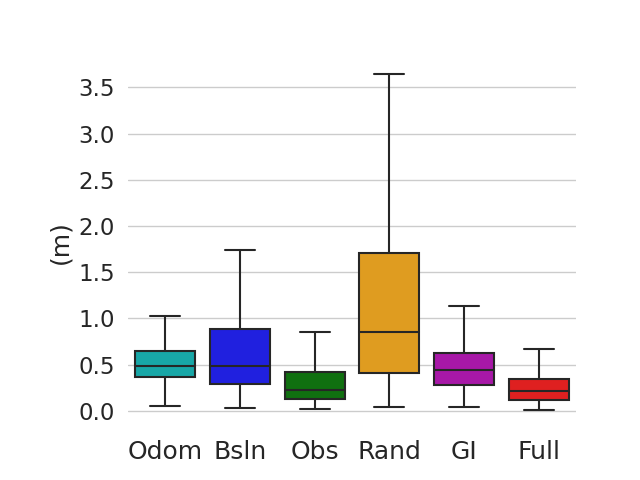}}
\hfill
\subfloat[Kentucky Underground]{\includegraphics[trim=20 0 30 20, clip, width=0.49\columnwidth]{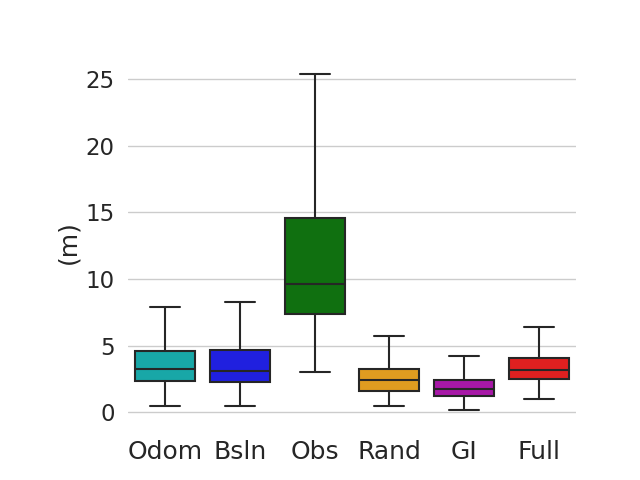}}
\caption{Trajectory error comparison on the powerful server on the four underground datasets. \rev{We find that the powerful server and less powerful computer (\cref{fig:ate-boxplots}) have similar performance when using our system.}}
 \vspace{-3mm}
\label{fig:ate-boxplots-ws}
\end{figure}

Finally, in Fig.~\ref{fig:ate-boxplots-ws} we provide another set of results by running 
the set of different configurations on the powerful server. Notice that the results obtained 
here is very similar to the set we obtained on the laptop,
which further emphasizes the benefit our proposed techniques have on the scalability challenge 
in loop closure detection.
\rev{We again find that the baseline and random solutions are outperformed by an odometry only solution in some situations. This is because there are many false positive loop closures generated which make the SLAM solution worse and these loop closures are not chosen by our prioritization methods as frequently.}

\subsection{RSSI Results}
\revfinal{We evaluate the RSSI results separately from the other modules due to its requirement of deploying radio beacons which were not deployed in the previous datsets.}
We evaluate the RSSI loop closure mechanism based on the the \textbf{Finals Dataset}, where the robot drops the beacons as described in \cite{Ginting2021}. 
Table \ref{tab:rssi_loop} shows the max and mean error for rotational and translational components for three robots: Spot1, Spot2, and Spot3.
The second column shows the proposed loop prioritization system with only the RSSI prioritization.
The third column presents the proposed system with Observability and Graph Information Prioritization. 
The result shows that RSSI-based loop closure gives similar or better results as the full system. 
That implies that additional loop generation mechanisms are not needed in some scenarios but only robust indicators of place recognition, such as provided by radio beacons.
Using RSSI-based prioritization is beneficial for robots if the computation resources are limited and loop closure detection is needed. 

\begin{table}[ht]
\caption{Evaluation of RSSI based loop closures.
 }
\label{tab:rssi_loop}
\centering
    \resizebox{1.0\columnwidth}{!}{
	\begin{tabular}{c | c c c  c  }
		\toprule
		Spot & Error & Odometry &  RSSI  & Full \\
		\midrule
		\multirow{2}{*}{1}
		& Mean (Max) Translational \% (m) & 1.52 (4.42)  &	\textbf{0.72} (\textbf{3.70}) &		0.93 (4.30)\\
		& Mean (Max) Rotational $\frac{\degree}{m}$ (\textdegree)& 46.4 ( 7.63) & 	\textbf{17.2} (\textbf{4.91})&	24.9  (8.64)\\
		\midrule
		\multirow{2}{*}{2}
		& Mean (Max) Translational \% (m) & 5.47 (7.93) &	\textbf{0.38} (\textbf{1.46}) &		0.44 (2.62) \\
		& Mean (Max) Rotational $\frac{\degree}{m}$ (\textdegree) & 97.7 (20.01) &	\textbf{8.8} (\textbf{3.33}) &	10.9 (3.92)\\
		\midrule
		\multirow{2}{*}{3}
		& Mean (Max) Translational\% (m) & 12.38 (4.39) &		0.53 (	2.39) &	\textbf{0.52} (\textbf{2.28}) \\
		& Mean (Max) Rotational $\frac{\degree}{m}$ (\textdegree) & 40.40 (13.39 ) &		12.7 (4.33) & 	\textbf{12.5} (\textbf{3.87})\\
	\end{tabular}
	}
	\vspace{-0.6cm}
	
\end{table}

\section{Conclusion}
Loop closure prioritization is a central problem to large-scale multi-robot SLAM as good loop closures allow for the creation of a large drift-free map.
In this paper, we tackled the problem of making loop closure detection more scalable for multi-robot systems
by providing techniques to prioritize loop closure candidates for computation.
We demonstrated our techniques in the LAMP system and 
showcased our results with challenging field datasets and demonstrating that,
with our techniques, we were able to improve the performance of the SLAM system by selecting better 
loop closures \rev{during mission execution}.
We also demonstrate that our results are largely invariant to the computational power of the central computing system for a multi-robot team, reinforcing how efficient the system is.

\vspace{-3mm}

\printbibliography 

\end{document}